# Hierarchical Deep Co-segmentation of Primary Objects in Aerial Videos

Jia Li[1,3]
Pengcheng Yuan[1]
Daxin Gu[1]
Yonghong Tian[2,3]

[1]State Key Laboratory of Virtual Reality Technology and Systems, School of Computer Science and Engineering, Beihang University

[2]National Engineering Laboratory for Video Technology, School of EE&CS, Peking University, Beijing, China.

[3]Pengcheng Laboratory, Shenzhen, China.

Primary object segmentation plays an important role in understanding videos generated by unmanned aerial vehicles. In this paper, we propose a large-scale dataset with 500 aerial videos and manually annotated primary objects. To the best of our knowledge, it is the largest dataset to date for primary object segmentation in aerial videos. From this dataset, we find most aerial videos contain large-scale scenes, small primary objects as well as consistently varying scales and viewpoints. Inspired by that, we propose a hierarchical deep co-segmentation approach that repeatedly divides a video into two sub-videos formed by the odd and even frames, respectively. In this manner, the primary objects shared by sub-videos can be co-segmented by training two-stream CNNs and finally refined within the neighborhood reversible flows. Experimental results show that our approach remarkably outperforms 17 state-of-the-art methods in segmenting primary objects in various types of aerial videos.

Recently, unmanned aerial vehicles (drones) have become very popular since it provides a new way to observe and explore the world. As a result, aerial videos generated by drones have been growing explosively. For these videos, one of the key tasks is to segment the primary objects, which can be used to facilitate subsequent tasks such as event understanding, scene reconstruction, drone navigation and visual tracking.

Hundreds of models have been proposed in the past decade to segment primary objects[15], which can be roughly divided into two categories. The first category contains image-based models that focus on detecting salient (primary) objects in images. In this category, classic models[1-4] focus on designing rules to pop-out salient targets and suppress distractors, while recent models[5-8] usually adopt the deep learning framework due to the availability of large-scale image datasets (e.g., the XPIE dataset[4]). The second category contains video-based models[16] that aim to segment a sequence of primary/foreground objects that consistently pop-out in the whole video. Similar to the image-based category, classic video-based models also design rules to segment primary objects by jointly considering the per-frame accuracy and inter-frame consistency[9]. Recently, with the presence of large-scale video datasets[17], several deep learning models[10, 11] have been proposed as well. In addition, some video object co-segmentation approaches[12,13] have been proposed as well to simultaneously segment a common category of objects from two or more videos.

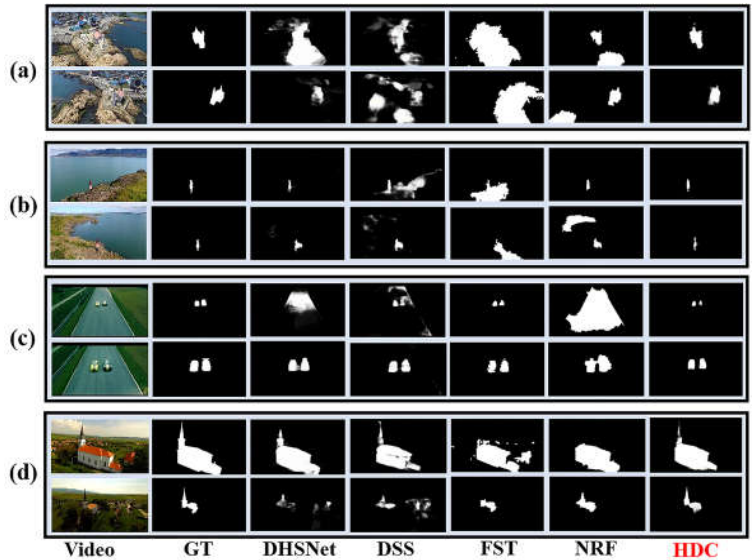

Figure 1: Representative challenging scenarios in aerial videos. (a) large-scale scenes, (b) small primary objects, (c) scale variation, (d) viewpoint variation. We also demonstrate the results of state-of-the-art models, including DHSNet[6], DSS[7], FST[9], NRF[11] and our approach (denoted as HDC).

Generally speaking, most existing models from the two categories can perform impressively on generic images and videos taken on the ground. However, their capability in processing aerial videos, which often contain large-scale scenes, small primary objects as well as consistently varying scales and viewpoints, may be not very satisfactory (see Figure 1 for some examples). The main reasons are two-folds: 1) the heuristic rules and learning frameworks may not perfectly fit the characteristics of aerial videos, and 2) there is a lack of large-scale aerial video datasets for model training and benchmarking. Toward this end, this paper proposes a large-scale dataset APD with 500 aerial videos (76, 221 frames). Based on the types of primary objects, these videos can be divided into five subsets, including humans, buildings, vehicles, boats and others. From these videos, 5,014 frames are sparsely sampled, in which the primary objects are manually annotated (see Figure 2 for representative frames and their ground-truth masks).

Based on the aerial video dataset APD, we propose a hierarchical deep co-segmentation approach for segmenting primary objects in aerial videos. In our approach, we first divide a long aerial video into two sub-videos formed by the odd and even frames, respectively. By repeatedly conducting such temporal slicing operations to the sub-videos, a long video can be represented by a set of hierarchically organized sub-videos. As a result, the object segmentation problem in a long aerial video can be resolved by hierarchically co-segmenting the objects shared by much shorter sub-videos. By learning end-to-end CNNs for co-segmenting two frames, a mask can be initialized for each frame by co-segmenting frames from sub-videos that have the same parent node in the hierarchy. These masks are then refined within the neighborhood reversible flows so that the primary video objects can consistently pop-out in the video. Experimental results show that our approach is efficient and outperforms 17 state-of-the-art models, including 7 image-based non-deep models, 5 image-based deep models and 5 video-based models. The results also show that APD is a very challenging dataset for existing object segmentation models.

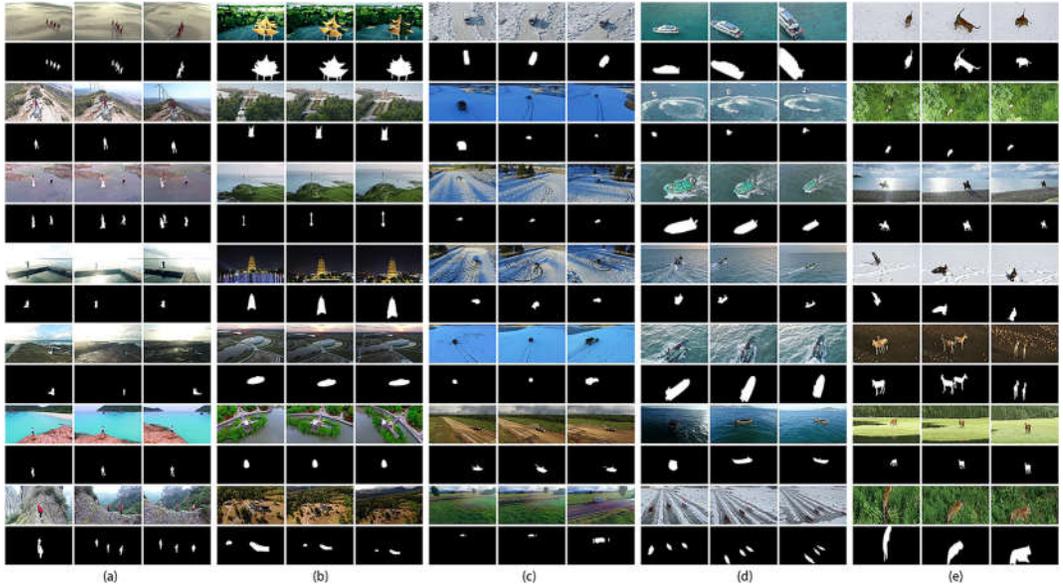

Figure 2: Frames and ground-truth masks from APD. (a) APD-Human (95 videos), (b) APD-Building (121 videos), (c) APD-Vehicle (56 videos), (d) APD-Boat (180 videos) and (e) APD-Other (48 videos).

The contributions are summarized as follows: 1) We propose a largest aerial video dataset for primary object segmentation, which can be used to further investigate the problem of primary video object segmentation from a completely new perspective; 2) we propose a deep co-segmentation approach that can efficiently and accurately segment primary objects in aerial videos; 3) we provide a benchmark of our approach and massive state-of-the-art models on the proposed dataset.

# APD: AN AERIAL SALIENCY DATASET

Towards primary object segmentation in aerial videos, we construct a large-scale dataset for model training and benchmarking, denoted as APD. In constructing the dataset, we first collect 2,402 long aerial videos (107 hours in total) shared on the Internet. Then we manually divide long videos into 52,712 shots and remove shots that are unlikely to be taken by drones or contain no obvious primary objects (determined through voting by three volunteers). After that, we obtain 21,395 video clips, from which we randomly sample 500 clips for the subsequent annotation process. According to the types of primary objects, these videos are further divided into five subsets, as shown in Table 1.

From these videos, we uniformly sample only one keyframe out of every 15 frames and manually annotate the 5,090 keyframes. In the annotation process, each annotator is requested to first watch the videos to obtain an initial impression of what are the primary video objects. Based on the impression, they then annotate the primary objects in the sparsely sampled keyframes with polygons. After that. the annotation quality of each frame is independently assessed by another two subjects. Flawed annotations are then corrected by the three annotators through majority voting, while frames with confusing annotations are discarded. Finally, we obtain 5,014 binary masks that indicate the location of primary video objects in keyframes.

Table 1: Dataset statistics. #Type: shooting from Ground or Aerial. #Max-F, #Min-F: the max and min numbers of frames. #Annot: the number of annotated frames. #Avg-Obj: the average number of objects per video or image. #Avg-Area: the average area of primary objects per video or image.

| Dataset | #Type | #Video | #Max Res | #Frames | #Max-F | #Min-F | #Annot | #Avg-Obj | #Avg-Area(%) |
|---|---|---|---|---|---|---|---|---|---|
| MS COCO2017 | G | | 640 × 640 | 158,791 | | | 157,403 | 2.51 ± 2.95 | 10.9 ± 0.20 |
| SegTrack V2 | G | 14 | 640 × 360 | 1,065 | 279 | 21 | 1,065 | 1.38 ± 1.01 | 7.38 ± 7.89 |
| ViSal | G | 17 | 512 × 288 | 963 | 100 | 30 | 193 | 1.16 ± 0.40 | 10.5 ± 6.51 |
| VOS | G | 200 | 800 × 800 | 116,103 | 2,249 | 71 | 7,467 | 1.15 ± 0.44 | 13.0 ± 12.6 |
| APD-Human | A | 95 | 720 × 1280 | 14,638 | 271 | 61 | 966 | 1.40 ± 0.64 | 1.50 ± 0.07 |
| APD-Building | A | 121 | 720 × 1280 | 17,749 | 271 | 31 | 1,170 | 1.16 ± 0.20 | 7.17 ± 0.57 |
| APD-Vehicle | A | 56 | 720 × 1280 | 7,665 | 280 | 31 | 505 | 1.14 ± 0.14 | 2.54 ± 0.09 |
| APD-Boat | A | 180 | 720 × 1280 | 28,085 | 280 | 16 | 1,851 | 1.27 ± 0.31 | 3.52 ± 0.32 |
| APD-Other | A | 48 | 720 × 1280 | 8,084 | 284 | 86 | 522 | 1.45 ± 0.58 | 2.80 ± 0.09 |
| APD | A | 500 | 720 × 1280 | 76,221 | 284 | 16 | 5,014 | 1.27 ± 0.36 | 3.86 ± 0.33 |

To demonstrate the major characteristics of APD, we show the statistics of APD and its subsets in Table 1. In addition, to facilitate the difference between APD and previous datasets, we also show the information of three representative datasets with ground-level videos for primary or salient object segmentation, including SegTrack V2[13], ViSal[14] and VOS[10] and a representative image dataset for object segmentation named MS COCO[18]. As shown in Table 1, the primary objects in APD are remarkably smaller than that in previous datasets. Such small objects will make the segmentation task of primary objects very difficult. Considering that there already exist many approaches for the detection, segmentation and recognition of humans and vehicles, the APD dataset provides an opportunity to find out a way that can transfer ground-level knowledge of humans and vehicles to aerial videos. Moreover, the number of videos in APD are larger than previous datasets, making ADP more diversity. In this sense, it is possible to directly train video-based deep learning models on APD with less risk of over-fitting.

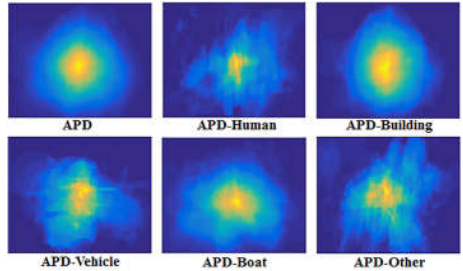

Figure 3: Average annotation maps of APD and its five subsets.

Beyond the quantitative statistics, we also show the average annotation maps of APD and its subsets in Figure 3. An average annotation map is computed by resizing all annotated masks to the same resolution, and normalizing the map to a maximum value of 1. From Figure 3, we find that the distribution of primary objects also has a strong center-bias tendency, implying that many rules and models for generic primary/salient object segmentation can be re-used for segmenting primary objects in aerial videos (e.g., the boundary prior[2]). Moreover, the degrees of center-bias in the five subsets differ from each other, indicating that there may exist several different ways to optimally segment primary objects in aerial videos if their semantic attributes are known or predictable.

# HIERARCHICAl DEEP CO-SEGMENTATION OF PRIMARY VIDEO OBJECTS

The segmentation task of primary objects in videos is to consistently pop-out the same primary object throughout the video. While the challenges of large-scale scenes, small objects and consistently varying scales and viewpoints make this task in aerial videos very challenging. Fortunately, we find that most primary objects last for a long period in the majority of aerial video sequences, which may be caused by the fact that aerial videos usually have less or slower camera motions and wider viewing angles. Inspired by this fact, we propose a novel approach for primary object segmentation in aerial videos by turning a complex task to several simple ones. The framework of our approach is shown in Figure 4, which consists of three major stages: 1) hierarchical temporal slicing of aerial videos, 2) mask initialization via video object co-segmentation and 3) mask refinement within neighborhood reversible flows. Details of these three stages are described as follows.

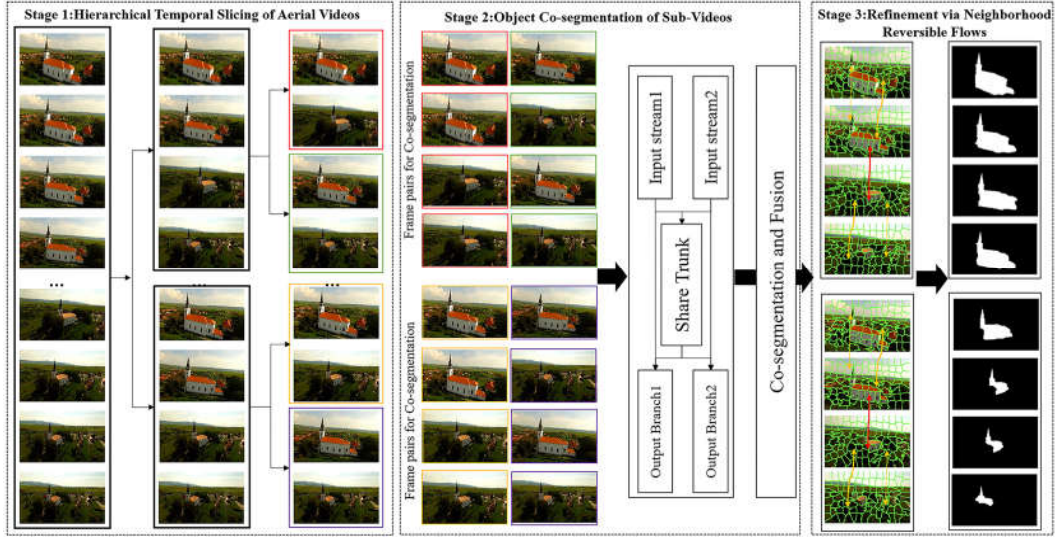

Figure 4: Framework of our approach.

## Hierarchical Video Slicing

In the first stage, we divide a long aerial video into two sub-videos formed by the odd and even frames, respectively. In this manner, the content similarity between these two sub-videos can be maximally guaranteed. By repeatedly conducting the odd-even slicing operations to all sub-videos, a hierarchy of short video clips can be efficiently constructed. Assuming that primary objects last for at least $N$ frames in an aerial video, we can build a tree structure with a depth of $\lfloor \log_2 N \rfloor$ and $2^{\lfloor \log_2 N \rfloor}$ nodes. Here we empirically set $N = \max(32, \text{video length})$. The short video clip at each leaf node has at least one frame that contains the primary objects. As a result, primary objects in the original video can be segmented by solving a set of simpler tasks: hierarchically co-segmenting the objects shared by massive much shorter video clips.

## Mask Initialization

In the second stage, we aim to initialize a mask of primary objects for each video frame by hierarchically co-segmenting the objects shared by the $2^{\lfloor \log_2 N \rfloor}$ short video clips at leaf nodes. To speed up this process, the co-segmentation is conducted only between two sub-videos that have the same parent node. Let $\mathbb{A} = \{A_i, i = 1, \cdots, |\mathbb{A}|\}$ and $\mathbb{B} = \{B_j, j = 1, \cdots, |\mathbb{B}|\}$ be two short video clips, where $|\mathbb{A}|$ and $|\mathbb{B}|$ denote the numbers of frames in $\mathbb{A}$ and $\mathbb{B}$, respectively. For these two short videos, we assume that there exists a model $\phi(A_i, B_j)$ that can segment the objects shared by the $i$th frame of $\mathbb{A}$ and the $j$th frame of $\mathbb{B}$:

$$\phi(A_i, B_j) = \{M_{A_i|B_j}, M_{B_j|A_i}\}, \quad (1)$$

where $M_{A_i|B_j}$ is a probability map for the frame $A_i$ that depicts the objects shared with the frame $B_j$. By co-segmenting all frame pairs between $\mathbb{A}$ and $\mathbb{B}$, the mask of primary objects for a frame $A_i$ can be initialized as the per-pixel average of all such co-segmentation results with respect to all frames from $\mathbb{B}$:

$$M_{A_i} = \frac{1}{|\mathbb{B}|} \sum_{j=1}^{|\mathbb{B}|} M_{A_i|B_j}. \quad (2)$$

From the map produced by (2), we find that a frame is actually co-segmented with multiple non-adjacent frames with increasing temporal distances. The advantages of such co-segmentation between far-away frames are at least four-folds: First, far-away frames can provide more useful cues of the primary objects in the co-segmentation process than adjacent frames that are full of redundant visual stimuli. In other words, far-away frames form a global picture of what is the primary video object. Second, most co-segmentation operations can pop-out primary objects since they appear in a large portion of video frames. As a result, primary objects can be repeatedly enhanced through the additive fusion in (2). Third, the hierarchical framework ensures that each frame can be co-segmented with at least one frame with the same primary objects. Last but not least, the computational cost of co-segmenting frame pairs from two short videos is remarkably smaller than that from two long videos so that the efficiency of the proposed approach can be improved.

In practice, the model $\phi(A_i, B_j)$ can be set to any co-segmentation algorithms. Here we train two-stream fully convolutional neural networks, denoted as CoSegNet, to conduct such co-segmentation. As shown in Figure 5, CoSegNet takes two frames as the input and two probability maps as the output. Features from the two frames are extracted with two separate streams, which are initialized with the architecture and parameters of the first several layers of ResNet-50. After that, the output features of these two streams are concatenated and fused into a shared trunk for extracting the common features of the two frames. Then the network splits into two separate branches that predict a probability map of shared objects for each input frame. Note that a skip connection from each input stream to the corresponding output branch is also used to regularize the generation of each probability map by introducing frame-specific low-level features.

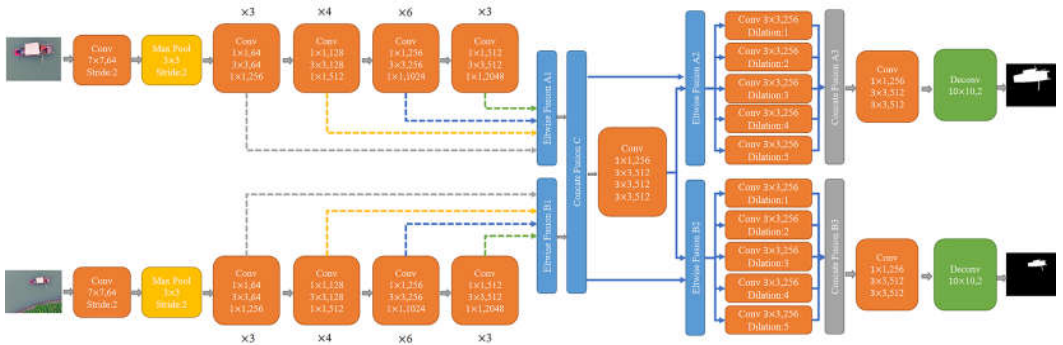

Figure 5: The network architecture of CoSegNet. The a × a, b inside each Conv block indicates kernel size $a$ and kernel number $b$, while the × c below each Conv block means $c$ sequential convolution layers in the Conv block.

For a pair of frames $I_i$ and $I_j$ with ground-truth masks $G_i$ and $G_j$, we train CoSegNet by simultaneously minimizing two losses $\mathcal{L}(M_{I_i|I_j}, G_i)$ and $\mathcal{L}(M_{I_j|I_i}, G_j)$, where $\mathcal{L}(\cdot)$ is the cross-entropy loss. We resize all input frames and output predictions to the resolution of 320×320. The learning rate is set to $1 \times 10^{-8}$ at the first 10 epochs and $2 \times 10^{-9}$ in subsequent iterations. A batch size of four frame pairs is adopted in training the network. The optimization algorithm is set to SGD, the gamma value is set to 0.2 and the momentum is set to 0.9.

In training CoSegNet, we utilize two types of data, including 1) synthetic data generated by randomly cropping a pair of 320×320 patches from an image with manually annotated salient objects (we use the same training images of[11] that are overwhelmed by ground-level scenarios), and 2) realistic data generated by randomly sampling pairs of annotated key-frames from the training set of APD. In this way, CoSegNet trained only on synthetic data is used as a baseline model to justify the effectiveness of our hierarchical deep co-segmentation framework, while CoSegNet trained on both synthetic and realistic data is used to give the state-of-the-art performance.

## Mask Refinement

After co-segmenting two short videos $\mathbb{A}$ and $\mathbb{B}$, each frame obtains an initial object mask represented by a probability map. Recall that the sub-videos $\mathbb{A}$ and $\mathbb{B}$ under the same parent node are generated by the odd and even frames of a longer sub-video $\mathbb{C} = \{C_1, C_2, \cdots, C_{|\mathbb{C}|}\}$, we assume each frame $C_u$ is initialized with a probability map $M_{C_u}$.

To enhance inter-frame consistency and correct probable errors in $M_{C_u}$, a key challenge is to derive reliable inter-frame correspondences. Considering that frames in the sub-video $\mathbb{C}$ may be actually far away from each other in the original video, the pixel-based optical flow may fail to handle large pixel displacement. To address this problem, we construct neighborhood reversible flows[11] based on superpixels. We first divide two frames $C_u$ and $C_v$ into $N_u$ and $N_v$ superpixels that are denoted as $\{\mathcal{O}_{ui}\}$ and $\{\mathcal{O}_{vj}\}$, respectively. Similar to[11], we compute the pair-wise $\ell_1$ distances between superpixels from $\{\mathcal{O}_{ui}\}$ and $\{\mathcal{O}_{vj}\}$, where a superpixel is represented by its average RGB, Lab and HSV colors as well as the horizontal and vertical positions. Suppose that $\mathcal{O}_{ui}$ and $\mathcal{O}_{vj}$ reside in the $k$ nearest neighbors of each other, they are $k$-nearest neighborhood reversible with the correspondence measured by

$$f_{ui,vj} = \begin{cases} \exp(-2k/\sigma) & k \leq \sigma \\ 0 & otherwise \end{cases} \quad (3)$$

where $\sigma$ is a constant empirically set to 15 to suppress the weak inter-frame correlations. Such superpixel-based inter-frame correspondence between $C_u$ and $C_v$ is denoted as the neighborhood reversible flow $\mathbf{F}_{u,v} \in \mathbb{R}^{N_u \times N_v}$, in which the component at $(i,j)$ equals $f_{ui,vj}$. Note that we further normalize $\mathbf{F}_{u,v}$ so that each row sums up to 1. Based on such flows, we refine the initial mask $M_{C_u}$ according its correlations with other frames. To speed up the refinement, we only refer to the previous mask $M_{C_{u-1}}$ and subsequent mask $M_{C_{u+1}}$. We first turn the pixel-based map $M_{C_u}$ to a vectorized superpixel-based map $\mathbf{x}_u$ by averaging the scores of all pixels inside each superpixel. After that, the score in $\mathbf{x}_u$ is updated as

$$\hat{\mathbf{x}}_u = \frac{\mathbf{x}_u + \lambda_p \cdot \mathbf{F}_{u,u-1} \mathbf{x}_{u-1} + \lambda_s \cdot \mathbf{F}_{u,u+1} \mathbf{x}_{u+1}}{1 + \lambda_p + \lambda_s}, \quad (4)$$

where $\lambda_p = \lambda_s = 0.5$ are two constants to balance the influence of previous and subsequent frames. After the temporal propagation, we turn superpixel-based scores into pixel-based ones as

$$\widehat{M}_{C_u}(p) = \sum_{i=1}^{N_u} \delta(p \in \mathcal{O}_{ui}) \cdot \hat{x}_{ui}, \quad (5)$$

where $\widehat{M}_{C_u}(p)$ is the refined probability map of the frame $C_u$ that depict the presence of primary objects at the pixel $p$. $\delta(p \in \mathcal{O}_{ui})$ is an indicator function which equals 1 if $p \in \mathcal{O}_{ui}$ and 0 otherwise. $\hat{x}_{ui}$ is the component in $\hat{\mathbf{x}}_u$ that corresponds to the superpixel $\mathcal{O}_{ui}$. An adaptive threshold $0.2 \times \max\{\widehat{M}_{C_u}(p), \forall p \in C_u\}$ is then used to segment the primary objects in the frame $C_u$.

# EXPERIMENTS

In the experiments, we compare our approach HDC with 17 state-of-the-art models on APD and VOS to demonstrate 1) the key challenges in APD, and 2) the effectiveness of the HDC. The models to be compared with can be divided into three groups:

1. The [I+N] group contains 7 image-based non-deep models, including DSR[15], MB+[11], GMR[15], SMD[1], RBD[2], HDCT[3] and ELE+[4].
2. The [I+D] group contains 5 image-based deep models, including RFCN[11], DCL[5], DHSNet[6], DSS[7] and FSN[8].
3. The [V] group contains 5 video-based models, including FST[9], SSA[10], NRF[11], MSG[12] and RMC[15].

In the comparisons, we divide APD into three subsets: 50% for training, 25% for validation and 25% for testing. The validation set is only used for parameter-finetuning and cannot be used to provide additional training data. On the testing subset with 125 videos, we evaluate the model performance with two metrics, including the mean Interaction-over-Union (mIoU) and the weighted F-Measure (wFM). The mIoU score is computed following the way proposed in VOS[10], which first computes the IoU score at each frame and then step-wisely average them on each video and the whole dataset. The thresholds for turning probability maps into binary masks are set to 20% of the maximal probability scores, as suggested in NRF[11]. Similarly, wFM is computed to assess the segmentation performance by jointly considering the completeness and exactness.

To show the challenges of APD, we list the model performance in Table 2 before fine-tuning them on APD. We find that APD is very challenging for most existing models. On this dataset, HDC outperforms the other models. The image-based non-deep models perform far from perfect, especially on the APD-Human subset since the primary objects cover only 1.5% area of the video frames on average. Most image-based deep models outperform non-deep ones, indicating that the learned features are more robust than heuristic rules when the application scenarios are transferred from ground-based to aerial. Moreover, NRF achieves impressive performance scores that are much higher than SSA and FST. This implies that the CNNs learned on ground-level image datasets can be partially reused in aerial videos, while the predictions can be further refined by using the inter-frame correspondences. Furthermore, the performances of some models, such as NRF and DHSNet, have different ranks in terms of mIoU and wFM. This phenomenon may imply that mIoU and wFM are two metrics that reveal the model performance from two different perspectives. Therefore, we suggest to use both metrics for model evaluation on APD.

To verify the generalization ability, we also test the models on VOS. From Table 2, we found that the performance of HDC is still the best, making it a scalable model that can be generalized to other scenarios.

Besides, for proving the efficiency of HDC, we test on the platform with a 3.4 GHz CPU (single core) and a NVIDIA GTX 1080 GPU (without batch processing). Note that we down-sample all videos to 320 × 320 for the fair comparison of various models in the speed test. As a result, we find that our approach takes only 0.73s to process a frame, which is much faster than almost all video-based models. Besides, the speed of HDC is comparable to many deep learning based models, such RFCN and DSS. The high efficiency of our approach makes it possible to be used in some real-world applications.

Table 2: Performance benchmark of HDC and state-of-the-art models before being fine-tuned on VOS and APD. The first two models are marked with bold and underline, respectively.

| | Models | APD-Human | | APD-Building | | APD-Vehicle | | APD-Boat | | APD-Other | | APD | | VOS | | Time |
|---|---|---|---|---|---|---|---|---|---|---|---|---|---|---|---|---|
| | | mIoU | wFM | mIoU | wFM | mIoU | wFM | mIoU | wFM | mIoU | wFM | mIoU | wFM | mIoU | wFM | |
| [N+I] | DSR | .149 | .243 | .250 | .334 | .219 | .310 | .232 | .318 | .327 | .425 | .222 | .329 | .480 | .533 | 4.03 |
| | MB+ | .115 | .189 | .219 | .276 | .241 | .325 | .247 | .313 | .315 | .406 | .220 | .300 | .534 | .583 | 0.02 |
| | GMR | .138 | .200 | .211 | .279 | .250 | .269 | .193 | .239 | .315 | .302 | .202 | .258 | .472 | .526 | 0.46 |
| | SMD | .244 | .304 | .240 | .312 | .306 | .354 | .329 | .376 | .416 | .481 | .294 | .365 | .530 | .589 | 0.89 |
| | RBD | .162 | .275 | .232 | .319 | .257 | .351 | .275 | .363 | .373 | .491 | .243 | .357 | .529 | .576 | 0.15 |
| | ELE+ | .268 | .295 | .330 | .360 | .425 | .453 | .411 | .451 | .521 | .452 | .371 | .417 | .532 | .505 | 7.80 |
| | HDCT | .137 | .288 | .241 | .342 | .234 | .413 | .249 | .413 | .314 | .542 | .221 | .396 | .509 | .567 | 3.35 |
| [I+D] | RFCN | .338 | .378 | .360 | .398 | .467 | .521 | .521 | .561 | .603 | .671 | .451 | .510 | .357 | .398 | 1.00 |
| | DHSNet | .394 | .472 | .387 | .438 | .523 | .601 | .572 | .655 | .626 | .701 | .493 | .581 | .707 | .755 | 0.03 |
| | DSS | .277 | .409 | .326 | .389 | .474 | .564 | .462 | .572 | .509 | .668 | .400 | .517 | .643 | .705 | 0.82 |
| | FSN | .286 | .324 | .363 | .393 | .507 | .554 | .519 | .580 | .589 | .664 | .443 | .505 | .682 | .730 | 0.08 |
| | DCL | .349 | .433 | .341 | .394 | .503 | .566 | .511 | .554 | .561 | .583 | .444 | .515 | .645 | .682 | 0.47 |
| [V] | SSA | .284 | .348 | .263 | .331 | .366 | .432 | .350 | .421 | .489 | .551 | .333 | .414 | .480 | .574 | 6.76 |
| | FST | .272 | .308 | .190 | .213 | .535 | .596 | .342 | .399 | .375 | .440 | .319 | .382 | .576 | .629 | 4.52 |
| | MSG | .080 | .088 | .160 | .186 | .216 | .232 | .143 | .172 | .361 | .410 | .153 | .182 | .374 | .420 | 14.3 |
| | RMC | .123 | .136 | .202 | .226 | .312 | .336 | .208 | .231 | .230 | .265 | .205 | .233 | .331 | .376 | 7.42 |
| | NRF | .393 | .433 | .423 | .449 | .507 | .540 | .552 | .598 | .677 | .741 | .496 | .551 | .711 | .761 | 0.18 |
| | HDC | **.462** | **.517** | **.515** | **.562** | **.649** | **.696** | **.619** | **.675** | **.726** | **.810** | **.582** | **.649** | **.739** | **.784** | 0.73 |

Beyond the direct performance comparisons, we fine-tune our HDC model and the other three top-performed deep models, DSS, NRF and DHSNet, on the training and validation sets of our APD dataset. The performance scores of the fine-tuned models (marked with *) are shown in Table 3. Some representative results of HDC* are shown in Figure 6.

From Table 3, we find that HDC* still performs much better than the other three deep models after all models are fine-tuned on APD. Although NRF DSS and DHSNet can learn some useful clues, they cannot deal with many aerial videos properly that the primary object is very small, especially these small objects are not always salient in all Frames. On the contrary, HDC* well resolves the problem from the perspective of co-segmentation. Even when the scene contains rich content and small-sized primary objects, the hierarchical co-segmentation framework can enforce HDC* to learn the features from the objects shared by a pair frames, leading to higher performance than single frame-based deep models.

Table 3: Performance comparison of HDC and the top three models after being fine-tuned on APD.

| Models | APD-Human | | APD-Building | | APD-Vehicle | | APD-Boat | | APD-Other | | APD | |
|---|---|---|---|---|---|---|---|---|---|---|---|---|
| | mIoU | wFM | mIoU | wFM | mIoU | wFM | mIoU | wFM | mIoU | wFM | mIoU | wFM |
| DHSNet* | .503 | .616 | .432 | .486 | <u>.713</u> | <u>.786</u> | .665 | <u>.759</u> | .687 | .754 | .587 | .685 |
| DSS* | <u>.524</u> | <u>.650</u> | .421 | .501 | .664 | .775 | .656 | .747 | .636 | .725 | .575 | <u>.688</u> |
| NRF* | .500 | .569 | <u>.492</u> | <u>.544</u> | .681 | .735 | <u>.679</u> | .744 | <u>.716</u> | <u>.799</u> | <u>.608</u> | .687 |
| HDC* | **.587** | **.668** | **.590** | **.665** | **.745** | **.801** | **.718** | **.786** | **.733** | **.820** | **.672** | **.758** |

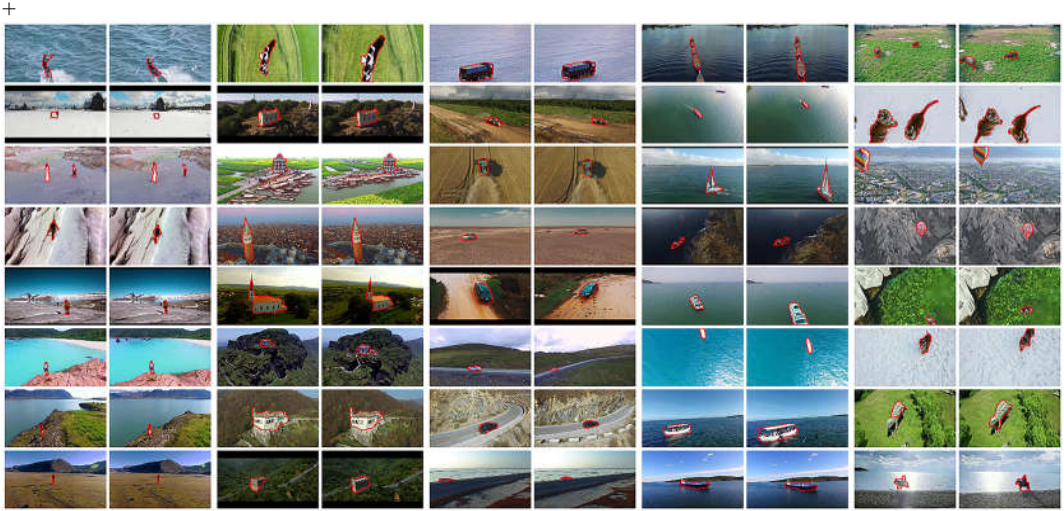

Figure 6: Representative results of HDC* on APD.

To validate the effectiveness of the odd-even temporal slicing framework, we test HDC* again by hierarchically dividing the testing videos into the same number of sub-videos formed by consecutive frames other than the even and odd frames. Note that the same HDC* model pretrained on the training set of APD is used for co-segmentation. In the case, the mIoU of HDC* decreases from 0.672 to 0.660, and the wFM decreases from 0.758 to 0.748, implying the odd-even slicing framework provides better frame pairs for co-segmentation.

In addition, we analyze the performance of HDC on APD before refinement to verify the influence of the refinement stage. We find that the mIOU and wFM drop to 0.563 and 0.639, respectively, which still achieve the highest performance compared with other previous methods. While due to the neighborhood reversible flow constructed in the mask refinement stage, we could further enhance inter-frame consistency and correct probable errors effectively.

Another concern may be the complexity and rationality of the hierarchical temporal slicing framework. By dividing testing videos into the depth 2, 3, 4, 5 and 6, we find that a deeper hierarchy leads to almost stable performance but remarkably lower complexity. For a video with 181 frames, the co-segmentation times are 785K, 379K, 177K, 76K and 25K when the depth is set to 2, 3, 4, 5 and 6. In most experiments, we adopt a depth of 5.

# CONCLUSION

This paper proposes a dataset, which is currently the largest, for primary object segmentation in aerial videos. We believe this dataset will be helpful for the development of video object segmentation techniques. Based on the dataset, we propose a hierarchical deep co-segmentation approach for primary video segmentation in aerial videos. The segmentation task is converted to a set of co-segmentation tasks that are easier to be resolved. Experimental results show that the proposed dataset is very challenging and the proposed approach outperforms 17 state-of-the-art models.

In the future work, we will try to explore the difference between the visual patterns extracted from ground-based and aerial videos so as to facilitate the design of better models for primary video object segmentation. In addition, the probability of constructing CNNs that can directly co-segment two short videos will be explored as well.

## ACKNOWLEDGMENTS

This work was supported in part by National Natural Science Foundation of China (61672072, 61532003 and 61825101), National Basic Research Program of China (2015CB351806) and Beijing Nova Program (Z181100006218063). Correspondence should be addressed to Yonghong Tian and Jia Li. Dataset and code can be found at http://cvteam.net

## ABOUT THE AUTHORS

**Jia Li** is an Associate Professor with the School of Computer Science and Engineering, Beihang University, Beijing, China. He received the B.E. degree from Tsinghua University in 2005 and the Ph.D. degree from the Institute of Computing Technology, Chinese Academy of Sciences, in 2011. His research interests include multimedia big data and learning-based visual content understanding. E-mail: jiali@buaa.edu.cn.

**Pengcheng Yuan** is pursuing the Master degree in the School of Computer Science and Engineering, Beihang University. His research interests include computer vision and machine learning. E-Mail: yuanpengcheng@buaa.edu.cn.

**Daxin Gu** is pursuing the Ph.D. degree in the School of Computer Science and Engineering, Beihang University. His research interests include computer vision and aerial video understanding. E-mail: gudaxin@buaa.edu.cn.

**Yonghong Tian** is a Full Professor with the National Engineering Laboratory for Video Technology, Peking University, Beijing, China. He received the Ph.D. degree from the Institute of Computing Technology, Chinese Academy of Sciences, in 2005. His research interests include computer vision and multimedia big data. He has been awarded two national prizes and three ministerial prizes in China. E-mail: yhtian@pku.edu.cn.